\documentclass{llncs}

\usepackage{epsfig}

\begin{document}

\title{An architecture for massively parallelization of the 
compact genetic algorithm}

\author{Fernando G. Lobo \and Cl\'audio F. Lima \and Hugo M\'artires}
\institute{ADEEC-FCT\\
   Universidade do Algarve\\
   Campus de Gambelas\\
   8000-062 Faro, Portugal\\
   \{flobo,clima\}@ualg.pt, hmartires@myrealbox.com 
}
\maketitle

\begin{abstract}
This paper presents an architecture which is suitable for a 
massive parallelization of the compact genetic algorithm. The
resulting scheme has three major advantages. First, it has 
low synchronization costs. Second, it is fault tolerant, and
third, it is scalable.

The paper argues that the benefits that can be obtained with
the proposed approach is potentially higher than those obtained 
with traditional parallel genetic algorithms.
In addition, the ideas suggested in the paper may also be 
relevant towards parallelizing more complex probabilistic 
model building genetic algorithms.
\end{abstract}

\section{Introduction}
\label{sec:introduction}

There has been several efforts in the field of evolutionary computation
towards improving the genetic algorithm's efficiency. One of 
the efficiency enhancement techniques that has been investigated, 
both in theory and in practice, is the topic of parallelization
\cite{Cantupaz:2000}.

In this paper, parallelization is investigated further, this time
in the context of Probabilistic Model Building Genetic Algorithms (PMBGAs),
a class of genetic algorithms whose operational mechanics differ
somewhat from those of the traditional GAs.

Efficiency is a very important factor in problem solving.
When talking about computer algorithms, efficiency is usually addressed
along two major axis: {\em Time} and {\em Memory} requirements needed
to solve a problem.
In the context of genetic and other evolutionary algorithms, there 
is another axis, {\em Solution Quality}, that also needs to be addressed. 
This third
aspect comes into play because many of the problems that genetic and 
evolutionary algorithms attempt to solve cannot be solved optimally
with 100\% confidence unless a complete enumeration of the search
space is performed. Therefore, genetic algorithms, as well
as many other methods, use a search bias to try to give good 
approximate solutions without doing a complete exploration of the 
search space.

Summarizing, efficiency in genetic algorithms translates into a 
3-objective problem: (1) maximize solution quality, (2) minimize
execution time, and (3) minimize memory resources. The latter is 
usually not a great concern because in traditional GAs, memory
requirements are constant throughout a run, leaving us with 
a tradeoff between solution quality and execution time.

The rest of the paper is organized as follows.
The next section presents background material on parallel GAs
and on PMBGAs.
Then, section~\ref{sec:motivation} raises some issues that 
can be explored when parallelizing PMBGAs that were not
possible with regular GAs. Section~\ref{sec:architecture}
presents an architecture that allows a massive parallelization
of the compact GA, and in section~\ref{sec:experiments}
computer experiments are conducted and its results are discussed. 
Finally, a number of extensions are outlined, and the paper finishes with 
a summary and some conclusions.

\section{Background}
\label{sec:background}

This section presents background material which is necessary for 
understanding the rest of the paper. It starts with a review of the 
major issues involved in parallelizing GAs, and then reviews the 
basic ideas of PMBGAs giving particular emphasis to the compact GA.

\subsection{Parallel GAs}
\label{sec:parallel_gas}

An important efficiency question that people are faced with in problem
solving is the following: Given a fixed computational time, what is the 
best way to allocate computer resources in order to have as good a 
solution as possible.

Under such a challenge, the idea of parallelization stands out naturally
as a way of improving the efficiency of the problem solving task. By using
multiple computers in parallel, there is an opportunity for delivering
better solutions in a shorter period of time.

Many computer algorithms are difficult to parallelize, but that is not
the case with GAs because GAs work with a population of solutions which 
can be evaluated independently of one another. Moreover, in many problems
most of the time is spent on evaluating solutions rather than on the
internal mechanisms of the GA operators themselves. Indeed, the time 
spent on the GA operators is usually negligible compared to the time spent 
on evaluating individual solutions.

Several researchers have investigated the topic of parallel GAs and the
major design issues are in choices such as using one or more populations,
and in the case of using multiple populations, decide when, with who, and
how often do individuals communicate with other individuals of other
populations.

Although implementing parallel genetic algorithms is relatively simple,
the answers to the questions raised above are not so straightforward and 
traditionally have only been answered by means of empirical experimentation. 
One exception to that has been the work of Cant\'u-Paz \cite{Cantupaz:2000}
who has built theoretical models that lead to rational decisions for 
setting the different parameters involved in parallelizing GAs. 
There are two major ways of implementing parallel GAs:

\begin{enumerate}
\item Using a single population.
\item Using multiple populations. 
\end{enumerate}

In single population parallel GAs, also called Master-Slave parallel GAs, 
one computer (the master) executes the GA operations and distributes 
individuals to be evaluated by other computers (the slaves). After
evaluating the individuals, the slaves return the results back to the
master. There can be significant benefits with such a scheme because
the slaves can work in parallel, independently of one another.
On the other hand, there is an extra overhead in communication costs
that must be paid in order to communicate individuals and fitness values
back and forth.

In multiple population parallel GAs, what would be a whole population 
in a regular non-parallel GA, becomes several smaller populations 
(usually called demes),
each of which is located in a different computer.
Each computer executes a regular GA and occasionally, individuals
may be exchanged with individuals from other populations. Multiple population
parallel GAs are much harder to design because there are more degrees of
freedom to explore. Specifically, four main things need to be chosen: (1) the 
size of each population, (2) the topology of the connection between the 
populations, (3) the number of individuals that are exchanged,
and (4) how often do the individuals exchange.

Cant\'u-Paz investigated both approaches and
concluded that for the case of the Master-Slave architecture, the benefits of 
parallelization occur mainly on problems with long function evaluation 
times because it needs constant communication. Multiple population
parallel GAs have less communication costs but do not avoid completely
the communication scalability problem. In other words, in either approach,
communication costs impose a limit on how fast parallel GAs can be.
To overcome this limitation, Cant\'u-Paz proposed a combination
of the two approaches in what was called {\em Hierarchical Parallel 
GAs}, and verified that when using such an approach it is possible 
to reduce the execution time more than by using either approach alone.
The interested reader is referred to the original source 
for the mathematical formulation and for additional information on the
design of parallel GAs.

\subsection{Probabilistic Model Building Genetic Algorithms}
\label{sec:pmbgas}

Probabilistic Model Building Genetic Algorithms (PMBGAs), also
referred by some authors as 
{\em Estimation of Distribution Algorithms} (EDAs), or
{\em Iterated Density Evolutionary Algorithms} (IDEAs),
are a class of Evolutionary Algorithms that replace the traditional 
variation operators, crossover and mutation, by the construction of a 
probabilistic model of the population and subsequent sampling from that 
model to obtain a new population of individuals. The 
operation of PMBGAs can be summarized by the following 5 steps:

\begin{enumerate}
\item Create a random population of individuals.
\item Apply selection to obtain a population of ``good'' individuals.
\item Build a probabilistic model of those good individuals.
\item Generate a new population according to the probabilistic model.
\item Return to step 2.
\end{enumerate}

Work on this area begun with simple probabilistic models that 
treated each gene independently, sometimes also called order-1 models.
Later, more complex algorithms were developed to allow dependencies
among genes. A detailed review of these algorithms can be found 
elsewhere \cite{Pelikan:02} \cite{larranaga:2001}.

The next subsection, reviews in detail the compact GA \cite{Harik:99e},
which is an example of an order-1 PMBGA, and whose parallelization is
discussed later in the paper.

\subsection{The Compact Genetic Algorithm}
\label{sec:cga}

Consider a 5-bit problem with a population of 10 individuals as shown
below:

\begin{center}
\begin{tabular}{|c c c c c|}
  \hline 
         1 &  0 &  0 &  0 & 0\\ 
         1 &  1 &  0 &  0 & 1\\
         0 &  1 &  1 &  1 & 1\\
         1 &  1 &  0 &  0 & 0\\
         0 &  1 &  1 &  0 & 1\\
         0 &  1 &  1 &  1 & 0\\
         1 &  1 &  0 &  0 & 0\\
         1 &  0 &  0 &  0 & 0\\  
         0 &  1 &  1 &  0 & 1\\  
         1 &  0 &  0 &  1 & 1\\  
  \hline
\end{tabular}
\end{center}

Under the compact GA, the population can be represented 
by the following probability vector:

\begin{center}
\begin{tabular}{|c|c|c|c|c|}
  \hline 
  0.6 & 0.7 & 0.4 & 0.3 & 0.5 \\ 
  \hline
\end{tabular}
\end{center}

The probabilities are the relative frequency counts of the number
of 1's for the different gene positions, and can be interpreted as
a compact representation of the population. In other words, the
individuals of the population could have been sampled from the 
probability vector.

Harik et al. \cite{Harik:99e} noticed that it was possible to mimic the behavior 
of a simple GA, without storing the population explicitly.
Such observation came from the fact that during the course of a regular 
GA run, alleles compete with each other at every gene position. At the 
beginning, scanning the 
population column-wise, we should expect to observe that 
roughly 50\% of the alleles have value 0 and  50\% of the alleles have 
value 1. As the search progresses, for each column, either the 
zeros take over the ones, or vice-versa. 
Harik et al. built an algorithm that explicitly simulates 
the random walk that takes place on the allele frequency makeup 
for every gene position.
The resulting algorithm, the compact GA, was shown to be operationally 
equivalent to a simple GA that does not assume any linkage between 
genes. 

The compact GA does not follow exactly the 5 steps mentioned previously
(in section~\ref{sec:pmbgas}) for a typical PMBGA, because the algorithm 
does not manipulate the population explicitly. Instead, it does so in an 
indirect way through the update step of $1/N$, where $N$ denotes the 
population size of a regular GA.

\section{Motivation for parallelizing PMBGAs}
\label{sec:motivation}

The main motivation for parallelizing PMBGAs is the same as the one
for parallelizing regular GAs, or any other algorithm: efficiency. 
By using multiple computers it is possible to make the algorithm run faster.

In many ways, parallelizing PMBGAs has many similarities with
parallelizing regular GAs. On the other hand, the mechanics of PMBGAs
are different from those of regular GAs, and it is possible to take advantage
of that. Specifically, it is possible to increase efficiency by exploring
the following two things:

\begin{enumerate}
\item Parallelize model building.
\item Communicate model rather than individuals.
\end{enumerate}

In regular GAs, the time spent on the GA operations (selection, crossover,
and mutation) is usually negligible compared to the time spent in fitness 
function evaluations. When using PMBGAs, and especially when
using multivariate models,
the model-building phase is much more compute intensive than the usual 
crossover and mutation operators of a regular GA. For many problems, such 
overhead can contribute to a significant fraction of the overall execution 
time. In such cases, it makes a lot of sense to parallelize the 
model-building phase. There has been a couple of research efforts 
addressing this topic \cite{Ocenasek:03} \cite{Lam:2002}.

Another aspect that makes PMBGAs very attractive for parallelization
comes from the observation that the model is a compact representation of 
the population, and it is possible to communicate the model rather than 
individuals themselves. Communication costs can be reduced this way 
because the model needs significant less storage than the whole population.
Since communication costs can be drastically reduced, it might make
sense to clone the model to several computers, and each computer could 
work independently on solving the problem by running a separate PMBGA.
Then, the different models would have to be consolidated (or mixed)
once in a while.
The next section presents an architecture that implements this idea 
with the compact GA.

\section{An architecture for building a massively parallel compact GA}
\label{sec:architecture}

This section presents an architecture which is suitable for a scalable
parallelization of the compact GA. Similar schemes can be done with other
order-1 PMBGAs. However, the connection that exists
between the population size and the update step, makes the compact GA
more suitable when working with very large populations, a topic that is
revisited later.

Since the model-building phase of the compact GA is trivial,
our study focuses only on the second item mentioned in 
section~\ref{sec:motivation}; communicate the model rather than
individuals. In the case of the compact GA, the model is represented 
by a probability vector of size $\ell$ ($\ell$ is the chromosome length). 
Each variable of the probability vector contains a value which has to 
be a member of a finite set of $N+1$ values ($N$ denotes the size of 
the population that the compact GA is simulating). The $N+1$ numbers 
correspond to all possible allele frequency counts for a particular
gene ($0$, $1$, $2$, \ldots, $N$), and can be stored with 
$\log_{2} (N+1)$ bits. Therefore, the probability vector can be represented
with $\ell \times \log_{2} (N+1)$ bits. This value is of a different 
order of magnitude than the $\ell \times N$ bits needed to represent a 
population in a regular GA, making it feasible to communicate the model 
back an forth between different computers. 

\begin{figure}
\centering
\epsfig{figure=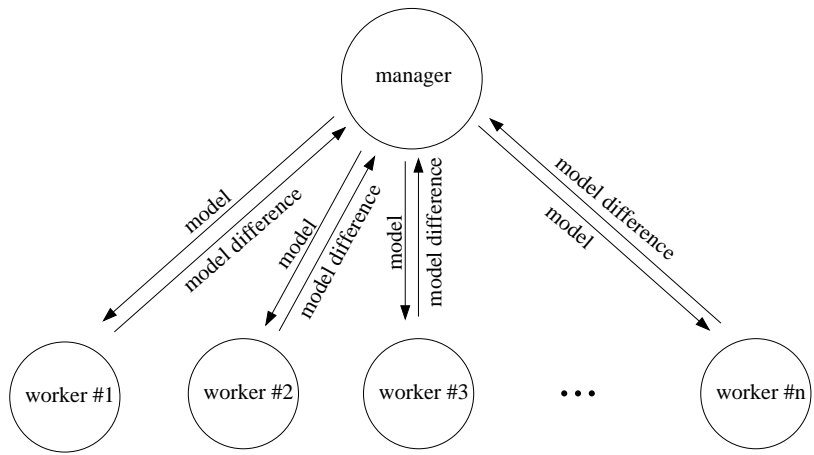,width=0.9\textwidth}
\caption{Manager-worker architecture.}
\label{fig:architecture}
\end{figure}

The storage savings are especially important when using large populations.
For instance, let us suppose that we are interested
in solving a 1000-bit problem using a population of size 1 million.
With a regular parallel GA, in order to communicate the whole population
it would be necessary to transmit approximately 1 Giga bit over a network.
Instead, with the compact GA, it would only
be necessary to transmit 20 thousand bits. The difference is large
and suggests that running multiple compact GAs in parallel with model
exchanges once in a while is something that deserves to be explored.
We have devised an architecture, that we call {\em manager-worker}, 
that implements this idea.
Figure~\ref{fig:architecture} shows a schematic of the approach.

Although Figure~\ref{fig:architecture} resembles
a master-slave configuration, we decided to give it a different name
(manager-worker) to contrast with the usual 
master-slave architecture of regular parallel GAs. There, the master 
executes and coordinates the GA operations and the slaves just compute 
fitness function evaluations. In the case of the 
parallel compact GA that we are suggesting, the manager also coordinates 
the work of the workers, but each worker runs a compact GA on its own.
There can be an arbitrary number of workers and there is no direct 
communication among them; the only communication that takes place 
occurs between the manager and a worker.

\subsection{Operational details}

One could think of different ways of parallelizing the compact GA. 
Indeed, some researchers have proposed different schemes \cite{Ahn:2003}
\cite{Hidalgo:2003}.
The
way that we are about to propose is particularly attractive because once the
manager starts, there can be an arbitrary number of workers, each of which
can start and finish at any given point in time making the whole system
fault tolerant. The operational details consist of the following seven steps:

\begin{enumerate}
\item The manager initializes a probability vector of size $\ell$ 
with each variable set to $0.5$. Then it goes to sleep, and waits to 
be woken up by some worker computer.

\item When a worker computer enters in action for the first time, 
it sends a signal to the manager saying that it is ready to start working.

\item The manager wakes up, sends a copy of its probability vector to 
the worker, and goes back to sleep. 

\item Once the worker receives the probability vector, it explores $m$ new
individuals with a compact GA. During this period, $m$ fitness function
evaluations are performed and the worker's local probability vector
(which initially is just a copy of the manager's probability vector)
is updated along the way.

\item After $m$ fitness function evaluations have elapsed,
the worker wakes up the manager in order to report the results
of those $m$ function evaluations. The results can be summarized by
sending only the differences that occurred between the vector that was
sent from the master and the worker's vector state after the execution
of the $m$ fitness function evaluations.

\item When the manager receives the probability vector differences sent 
by the worker, it updates its own probability vector by adding the differences
to its current vector. 

\item Then it sends the newly updated probability vector back to the worker.
The manager goes back to sleep and the worker starts working for $m$
more fitness function evaluations (back to step 4).
\end{enumerate}

There are a number of subtle points that are worth mentioning.
First of all, step number 7 is not a broadcast operation. The manager
just sends its newly updated probability vector to one particular worker. 
Notice however, that the manager's probability vector not only incorporates
the results of the $m$ function evaluations performed by that particular
worker, but it also incorporates the results of the evaluations conducted
by the other workers. That is, while a particular worker is working, other
workers might be updating the manager's probability vector. Thus, at a given
point in time, workers are working with a slightly outdated probability
vector. Although this might seem a disadvantage at first sight, the error
that is committed by working with a slightly outdated probability vector is
likely to be negligible for the overall search because an iteration
of the compact GA represents only a small step in the action of the GA
(this is especially true for large population sizes). 
The proposed parallelization scheme has several advantages, namely:

\begin{itemize}
\item Low synchronization costs.
\item Fault tolerance.
\item Scalability.
\end{itemize}

All the communication that takes place consist of short transactions. 
Workers do their job independently and only interrupt the manager once 
in a while. During the interruption period, the manager communicates
with a single worker, and the other workers can continue working non-stop.

The architecture is fault tolerant because workers can go up or down at any
given point in time. This makes it suitable for massively parallelization
using the Internet. It is scalable because potentially there is no limit 
on the number of workers.

\section{Computer simulations}
\label{sec:experiments}

This section presents computer simulations that were done to validate
the proposed approach. For the purpose of this paper, we are only interested 
in checking if the idea is valid. Therefore, and in order to simplify 
both the implementation and the interpretation of the results, we decided to 
do a serial implementation of the parallel compact GA architecture.
Although it might seem strange (after all, we are describing a scheme
for doing massive parallelization), doing a serial simulation of the behavior
of the algorithm has a number of advantages:

\begin{itemize}
\item we can analyze the algorithm's behavior under careful controlled
conditions.
\item we can do scalability tests by simulating a parallel compact GA with
a large number of computers without having the hardware.
\item we can ignore network delays and different execution speeds of different
machines.
\end{itemize}

The serial implementation that we have developed simulates that there
are a number of $P$ worker processors and 1 manager processor. The 
$P$ worker processors start running at the same time and they all 
execute at the same speed. In addition, it is assumed that the communication
cost associated with a manager-worker transaction takes a constant time 
which is proportional to the probability vector's size. 
Such a scheme can be implemented by having a collection of $P$
regular compact GAs, each one with its own probability vector, and iterating
through all of them, doing a small step of the compact GA main loop, 
one at a time. After a particular compact GA worker completes $m$ fitness function
evaluations, the worker-manager communication is simulated as illustrated
during section~\ref{sec:architecture}.

We present experiments on a single problem, a bounded deceptive
function consisting of the concatenation of 10 copies of a 3-bit trap
function with deceptive-to-optimal ratio of $0.7$ \cite{Deb:93a*}.
This same function has been used in the original compact GA work.
We simulate a selection rate of $s=8$
and did tests with a population size of $N=100000$ individuals
(each worker processor runs a compact GA that simulates
a 100000 population size). We chose this population size because we
wanted to use a size large enough to solve all the building blocks
correctly. We use $s=8$ following the recommendation given by Harik et al.
in the original compact GA paper for this type of problem. Finally, we
chose this problem as a test function because, even though the compact
GA is a poor algorithm in solving the problem, we wanted to use a function
that requires a large population size because those are the situations where
the benefits from parallelization are more pronounced.

Having fixed both the population size
and the selection rate, we decided to systematically vary the number
of worker processors $P$, as well as the $m$ parameter which has an 
effect on the rate of communication that occurs between the
manager and a worker.
We did experiments for $P$ in \{1, 2, 4, 8, 16, 32, 64, 128, 256, 512, 1024\},
and for a particular $P$, we varied the parameter $m$ in 
\{8, 80, 800, 8000, 80000\}. This totalled 55 different configurations,
each of which was run 30 independent times.

\begin{figure}[htb]
\centering
\mbox{
    {\epsfig{figure=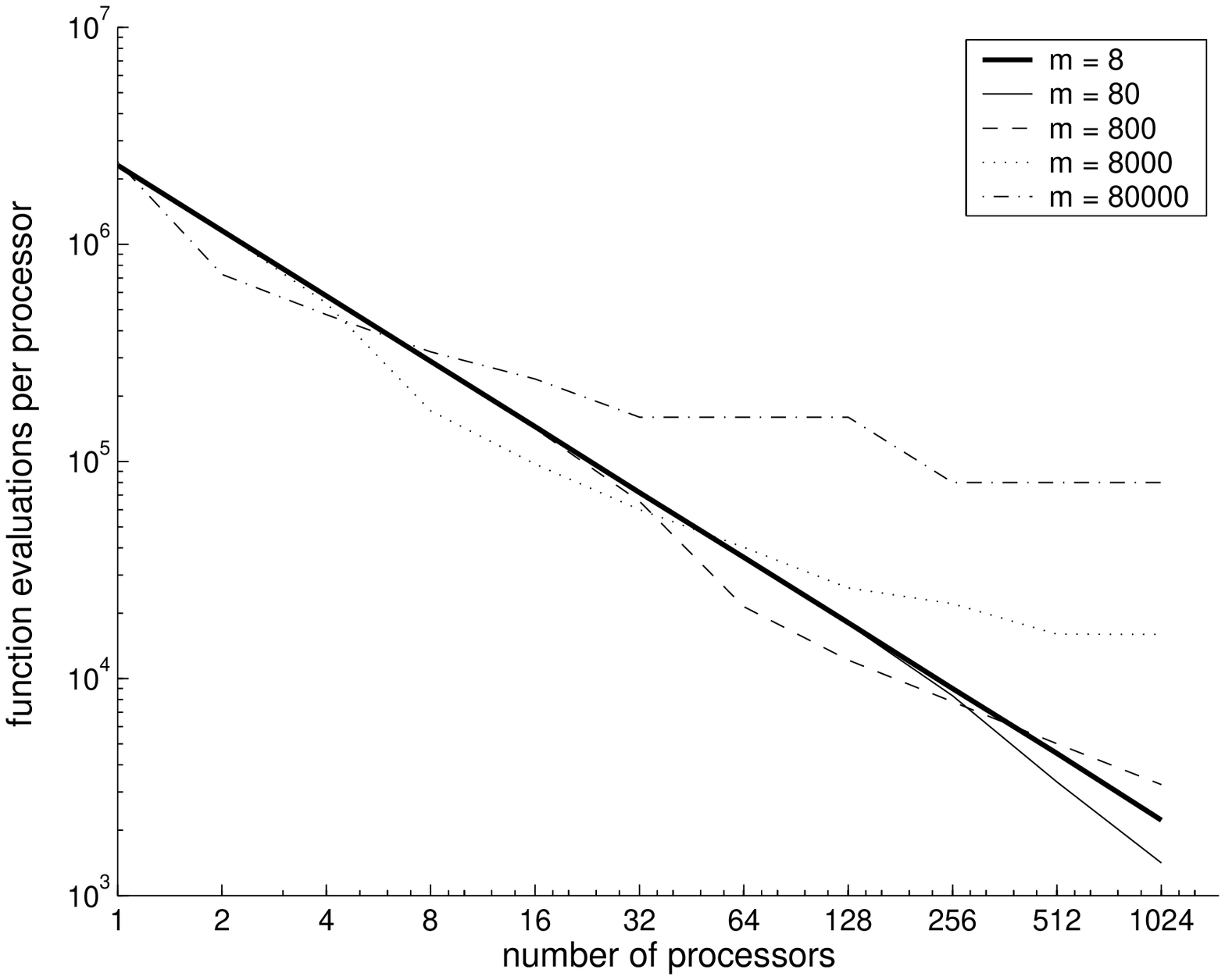, width=.48
        \textwidth}}\quad
    {\epsfig{figure=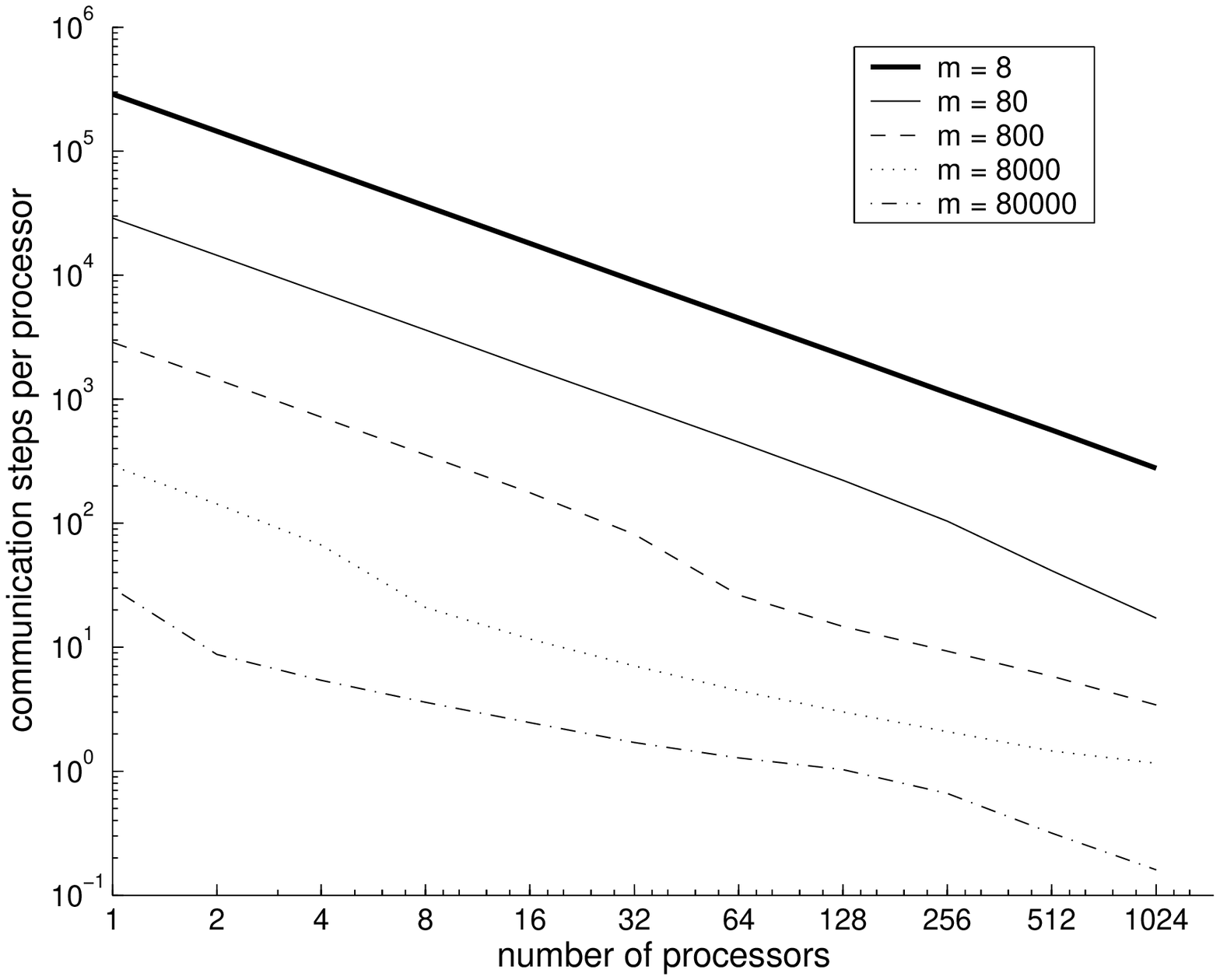, width=.48
        \textwidth}}
}
\caption{Both graphs depict a log-log plot. On the left, we see the average number of
function evaluations per processor. On the right, we see the average number of communication
steps per processor.}
\label{fig:experiments}
\end{figure}

The $m$ parameter is important because it is the one that affects communication
costs. Smaller $m$ values imply an increase in communication costs. 
On the other hand, for very large $m$ values, performance degrades because
the compact GA workers start sampling individuals from outdated probability
vectors.

Figure~\ref{fig:experiments} shows the results. In terms of fitness function
evaluations per processor, we observe a linear speedup for low $m$ values. 
For instance, for $m=8$
we observe a straight line on the log-log plot. Using the data directly, we
calculated the slope of the line and obtained an approximate value of -0.3.
In order to take into account the different logarithm bases, we need to multiply
it by $\log_{2}10$ (y-axis is $\log_{10}$, x-axis is $\log_{2}$) yielding a 
slope of approximately -1.
This means that the number of function evaluation per processor decreases
linearly with a growing number of processors.
That is, whenever we
double the number of processors, the average number of fitness function
evaluations per processor gets cut by a half. 

Likewise, in terms of communication costs, as we raise the parameter $m$,
the average number of communication steps between manager and worker decreases
in the same proportion as expected. For instance, for $m=80$, communication
costs are reduced 10 times when compared with $m=8$. 
Notice that there is a degradation in terms of speedup for the larger $m$ values.
For instance, 
for $m=8000$ and $m=80000$ (which is about the same order of the population size),
the speedup obtained goes away from the idealized case. This can be explained 
by the fact that in this case (and 
especially with a large number of processors), the average number of communication
steps per processor approaches zero. That means that a large fraction of processors
were actually doing some work but never communicated their results back to the
manager because the problem was solved before they had a chance to do so.

\section{Extensions}
\label{sec:extensions}

This work has a number of extensions worthwhile exploring.
Below, we outline some of them:

\begin{itemize}
\item Build theory for analyzing the effect of $m$, $N$, and $P$.
\item Compare with traditional parallel GA schemes.
\item Extend the approach to multivariate PMBGAs.
\item Take advantage of the Internet and build something like SETI@@home.
\end{itemize}

It would be interesting to study the mathematical analysis
of the proposed parallel compact GA. A number of questions come
to mind. For instance, what is the effect of the $m$ parameter?
What about the number of workers $P$? Should $m$ be adjusted
automatically as a function of $P$ and $N$? Our experiments suggest
that there is an ``optimal'' $m$ that depends on the number of 
compact GA workers $P$, and most likely depends on the 
population size $N$ as well.

Another extension that could be done is to compare
the proposed parallel architecture with those 
used more often in traditional parallel GAs, either master-slave
and multiple deme GAs. Again, our experiments suggest that the
parallel compact GA is likely to be on top of regular parallel
GAs due to lower communication costs. 

The model structure of the compact GA never changes, every gene is 
always treated independently. There are other PMBGAs 
that are able to learn a more complex structure dynamically as
the search progresses. One could think of using some of the ideas
presented here for parallelizing these more complex PMBGAs. 

Finally, it would be interesting to have a parallel compact GA implementation
based on the Internet infrastructure, where computers around the world could 
contribute with some processing power when they are idle. Similar schemes have
been done with other  projects, one of the most well known is
the  SETI@@home project \cite{Korpela:2001}. Our parallel GA architecture
is suitable for a similar kind of project because computers can go up or
down at any given point in time.

\section{Summary and conclusions}
\label{sec:summary_conclusions}

This paper reviewed the compact GA and presented an architecture that 
allows its massive parallelization. The motivation for doing so has
been discussed and a serial implementation of the parallel architecture
was simulated. Computer experiments were done under idealized conditions 
and we have verified an almost linear speedup with a growing number
of processors.

The paper presented a novel way of parallelizing GAs. This was possible
due to the different operational mechanisms of the compact GA when compared
with a more traditional GA. By taking advantage of the compact representation 
of the population, it becomes possible do distribute its representation to 
different computers without the associated cost of sending it individual
by individual.

Additional empirical and theoretical research needs to be done to confirm
our preliminary results. Nonetheless, the speedups observed in our experiments 
suggest that a massive parallelization of the compact GA may constitute an efficient
and practical alternative for solving a variety of problems.



\bibliographystyle{splncs}
\bibliography{references}

\end{document}